\documentclass[letterpaper]{article} 
\usepackage{aaai23}  
\usepackage{times}  
\usepackage{helvet}  
\usepackage{courier}  
\usepackage[hyphens]{url}  
\usepackage{graphicx} 
\usepackage{natbib}  
\usepackage{caption} 
\usepackage{algorithm}
\usepackage{algorithmic}

\usepackage{newfloat}
\usepackage{listings}
\DeclareCaptionStyle{ruled}{labelfont=normalfont,labelsep=colon,strut=off} 
\floatstyle{ruled}
\newfloat{listing}{tb}{lst}{}
\floatname{listing}{Listing}
\usepackage{tikz}
\usepackage{comment}
\usepackage{amsmath,amssymb} 
\usepackage{color}
\usepackage{graphicx}
\usepackage{amsmath}
\usepackage{amssymb}
\usepackage{booktabs}
\usepackage{bm}
\usepackage{multirow}
\usepackage{caption}
\usepackage{subcaption}

\newcommand{\RR}{\mathbb{R}}

\newcommand{\decoder}{\text{Dec}}
\newcommand{\encoder}{\text{Enc}}

\newcommand{\quantizedcode}{z_{\mathbf{q}}}
\DeclareMathOperator*{\argmin}{arg\,min}

\newcommand{\eg}{\textit{e.g. }}
\newcommand{\ie}{\textit{i.e. }}

\usepackage[capitalize]{cleveref}
\crefname{section}{Sec.}{Secs.}
\Crefname{section}{Section}{Sections}
\Crefname{table}{Table}{Tables}
\crefname{table}{Tab.}{Tabs.}

\title{PeCo: Perceptual Codebook for BERT Pre-training of Vision Transformers}
\author{Xiaoyi Dong$^{1}$\thanks{Equal contribution, $\dagger$ Corresponding Author}  \thanks{Work done during an internship at Microsoft Research Asia}, Jianmin Bao$^{2*}$,  Ting Zhang$^{2}$,  Dongdong Chen$^{3,\dagger}$,  Weiming Zhang$^{1}$ \\   Lu Yuan$^{3}$,  Dong Chen$^{2}$, Fang Wen$^{2}$,   Nenghai Yu$^{1}$, Baining Guo$^{2}$\\}
\affiliations{
    \textsuperscript{\rm 1}University of Science and Technology of China\\
    \textsuperscript{\rm 2}Microsoft Research Asia \ \textsuperscript{\rm 3}Microsoft Cloud + AI\\
    \{dlight@mail., zhangwm@, ynh@\}.ustc.edu.cn \ cddlyf@gmail.com \\
    \{jianbao, ting.zhang, luyuan, doch, fangwen, bainguo \}@microsoft.com
 
}

\begin{document}

\maketitle

\begin{abstract}
This paper explores a better prediction target for  BERT pre-training of vision transformers. 
We observe that current prediction targets disagree with human perception judgment.
This contradiction motivates us to learn a perceptual prediction target.
We argue that perceptually similar images should stay close to each other in the prediction target space. 
We surprisingly find one simple yet effective idea: enforcing perceptual similarity during the dVAE training. 
Moreover, we adopt a self-supervised transformer model for deep feature extraction and show that it works well for calculating perceptual similarity.
We demonstrate that such learned visual tokens indeed exhibit better semantic meanings, and help pre-training achieve superior transfer performance in various downstream tasks. For example, we achieve $\textbf{84.5\%}$ Top-1 accuracy on ImageNet-1K with ViT-B backbone, outperforming the competitive method BEiT by $\textbf{+1.3\%}$ under the same pre-training epochs. Our approach also gets significant improvement on object detection and segmentation on COCO and semantic segmentation on ADE20K. Equipped with a larger backbone ViT-H, we achieve the state-of-the-art ImageNet accuracy (\textbf{88.3\%}) among methods using only ImageNet-1K data.
\end{abstract}
\section{Introduction}

\label{sec:intro}

Current state-of-the-art self-supervised pre-training methods~\cite{dosovitskiy2020image,bao2021beit,he2021masked,xie2021simmim,chen2022context,wei2021masked} for vision transformers focus on masked image modeling (MIM),
a task of making predictions for masked patches from the visible patches. The input is usually an image consisting of visible patches and randomly masked patches and each patch is associated with corresponding positional embedding.
The prediction target for masked patches varies for different methods,
ranging from pixel-level prediction~\cite{dosovitskiy2020image,he2021masked,xie2021simmim} to feature-level prediction~\cite{bao2021beit,chen2022context,wei2021masked}.
In this paper, we study the prediction targets and introduce a better prediction target for MIM. 

\begin{figure}[t]\centering
\includegraphics[width=0.995\linewidth]{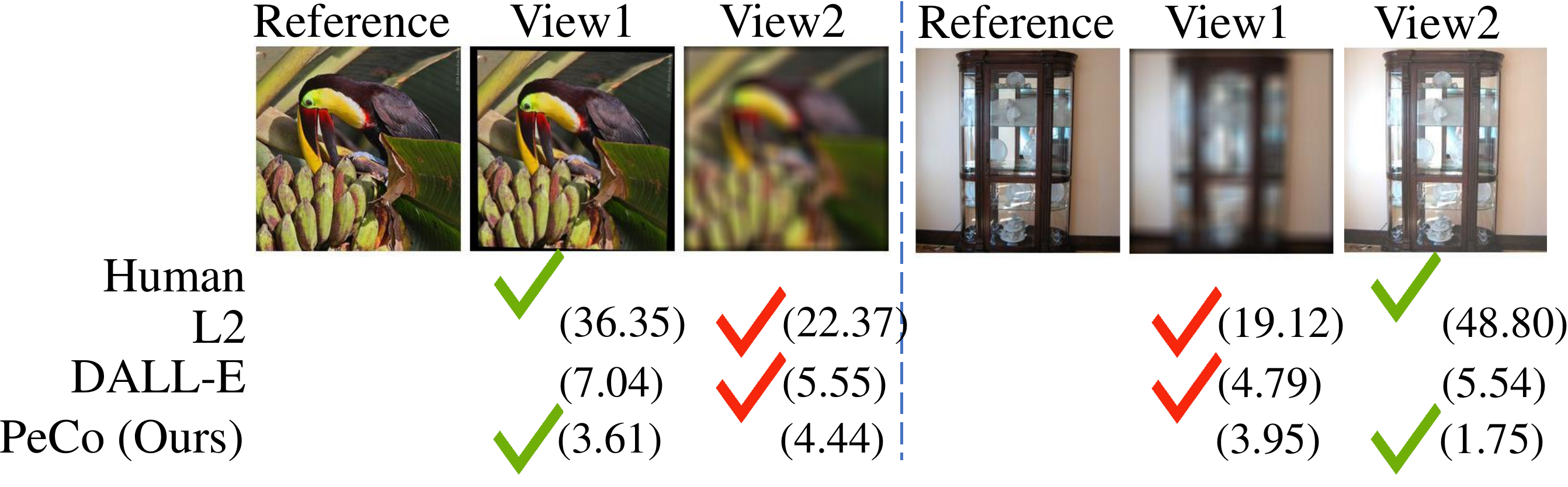}

\caption{Several examples illustrating the results of different prediction targets on the question that \textbf{which image (View1 or View2) is ``closer" to the Reference image}. The number denotes the distance between View1 or View2 and the Reference image. The images with smaller distances are considered more similar. We observe that the proposed PeCo agrees with human judgments while L2 or DALL-E disagree.}
\label{fig:perceptual}
\end{figure}

We point out that current prediction targets disagree with human judgment when evaluating the similarity between two different images.
There are two representative prediction targets in current MIM methods:
per-pixel regression and discrete token prediction.
Figure~\ref{fig:perceptual} illustrates the results of different prediction targets on the question that which image (View1 or View2) is ``closer" to the ``Reference" for these examples.
The reason for such disagreement of current prediction targets may come from 
the per-pixel loss. Note that the discrete tokens are obtained by a VQ-VAE trained under the objective of reconstruction loss, \ie, per-pixel loss.
The per-pixel measure assuming pixel-wise independence is insufficient for assessing structured outputs. For example, blurring causes large perceptual change but small pixel error, while shifting incurs small perceptual change but large pixel error~\cite{zhang2018unreasonable}.
Such disagreement with human visual perception indicates that perceptually similar patches may have divergent prediction targets.
This undermines the capability of MIM as it, in principle, is based on context prediction.

We propose that a good prediction target for MIM should coincide with human judgment. In other words, perceptually similar images should be close to each other in the prediction target space. Inspired from the observation in~\cite{zhang2018unreasonable} that deep features model low-level perceptual similarity surprisingly well, we introduce this so-called perceptual loss in VQ-VAE for discrete token learning. This loss can be viewed as per-feature loss as it aims to minimize the feature-wise distance between the original image and the reconstructed image. Specifically,
we adopt multi-scale deep features from multiple layers at different depth of a self-supervised Transformer. As shown in Figure~\ref{fig:perceptual}, our proposed new prediction target indeed aligns with human perception judgment.
We also show that the proposed visual tokens get much higher linear accuracy than the one without the perceptual loss. It indicates that our new visual tokens exhibit more semantic meanings, which is analogous to texts whose discrete tokens often contain highly semantic information.

We denote MIM using the introduced perceptual visual tokens for targets as ``PeCo", \ie Perceptual Codebook for BERT pre-training of vision transformers.
In the experiments, we demonstrate that equipped with such perceptual visual tokens, PeCo achieves better performance compared with the strong competitor BEiT~\cite{bao2021beit} using DALL-E~\cite{ramesh2021zero} codebook trained over $250M$ images without the perceptual loss.
We fine-tune the pre-trained model on various downstream tasks: image classification, object detection, and semantic segmentation. Experimental results show that our pre-trained model transfers better than  BEiT with only the prediction target changed. Concretely, we achieve $\textbf{84.5}\%$ Top-1 accuracy on ImageNet-$1$K with ViT-B model, outperforming BEiT by $\textbf{+1.3\%}$ with the same 800 pre-training epochs. Our approach also gets significant improvement on COCO object detection and semantic segmentation
as well as on ADE20K semantic segmentation. Our PeCo also shows strong scalability that when equipped with a larger backbone ViT-H, we achieve the state-of-the-art ImageNet accuracy (\textbf{88.3\%}) among methods using only ImageNet-1K data.

\section{Related Works}

\noindent \textbf{Self-supervised Learning.}
Self-supervised learning has attracted increasing attention over the past few years,
as deep learning networks become more and more data-hungry and it's impossible to label everything in the world.
There are two main categories along this path, contrastive and generative~\cite{liu2021self}.
One emerging field is self-supervised contrastive learning,
training an encoder to the representation measured by contrastive loss~\cite{hadsell2006dimensionality,dosovitskiy2014discriminative} via comparing similar and dissimilar samples. The representative methods include MOCO~\cite{he2020momentum,chen2020improved}, SimCLR~\cite{chen2020simple,chen2020big}, BYOL~\cite{grill2020bootstrap}, SwAV~\cite{caron2020unsupervised} and more~\cite{oord2018representation,li2021improve,bachman2019learning}.
However, contrastive-based methods heavily depend on the strong data augmentation and effective negative sampling.

The other recent resurgent field is generative self-supervised learning, 
training an encoder and a decoder under the objective of reconstruction loss.
The typical 
objectives, autoregressive and denoising autoencoder, aiming at recovering the corrupted or masked input, has yielded the most successful frameworks~\cite{devlin2018bert,radford2018improving,radford2019language,brown2020language,liu2019roberta,joshi2020spanbert} in NLP. 
Thanks to the pre-existing vocabulary in language, recovering the missing word can be transformed into
predicting all the possible words with the probability estimation, converting the prediction problem to an easier classification problem.
While in CV, on the other hand, most attempts~\cite{van2016pixel,oord2016conditional,chen2020generative,he2021masked} still resort to regression for generative methods
due to the lack of visual vocabulary, 
\eg iGPT~\cite{chen2020generative}. 
Recently, BEiT~\cite{bao2021beit} successfully adopts a classifier for prediction by directly adopting a VQ-VAE as the visual tokenizer.
Yet there exists a major difference between the language vocabulary and the visual vocabulary.
That is, the words of language are highly semantic, while the visual words of images are mostly not.
Most recently, numerous works~\cite{bao2021beit,he2021masked,xie2021simmim,chen2022context,wang2022bevt,dong2022bootstrapped, baevski2022data2vec,zheng2022general} based on MIM have been concurrently developed,
yet few studied the perceptual level of the prediction targets.
In this work, we attempt to learn a perceptual visual vocabulary for BERT pre-training, showing superior transfer performance than BEiT~\cite{bao2021beit} and MAE~\cite{he2021masked}.

\noindent \textbf{Discrete Visual Supervision.}
Exploring masked image modeling or image inpainting task for self-supervised pretrained tasks has never been stopped in vision community, especially when BERT~\cite{devlin2018bert} achieves great success in various tasks of NLP. To apply the cross-entropy loss function for vision tasks, iGPT~\cite{chen2020generative} clusters the pixel values to simulate the process of BPE~\cite{sennrich2015neural} process for different words in language. ViT~\cite{dosovitskiy2020image} attempts to directly divide the raw pixel values into multiple groups and assign a discrete label for each group GRB value. Recent work VQ-VAE~\cite{oord2017neural} proposes to adopt encoder and decoder to quantize the visual contents to a learnable codebook with fixed size.

\noindent \textbf{Perceptual Similarity.}
The perceptual similarity, as its name suggests, is to mimic the human perceptual
judgment of image similarity.
Numerous efforts have been proposed to achieve that, such as SSIM~\cite{wang2004image}, MSSIM~\cite{wang2003multiscale}, FSIM ~\cite{zhang2011fsim}, and HDR-VDP~\cite{mantiuk2011hdr}.
It has been shown in~\cite{zhang2018unreasonable} that the internal activations of network trained for classification task surprisingly coincide with human judgment. 
Such deep features have been widely used in image generation~\cite{gatys2016image,johnson2016perceptual,chen2017stylebank,bruna2015super,ledig2017photo,esser2021taming} with the goal of synthesizing realistic images.
The loss is called perceptual loss or VGG loss as the network used is often VGG architecture.
In this paper, we surprisingly discover that this simple loss is super effective in building a better prediction target and significantly improves vision BERT pretraining. 
Moreover, to enable self-supervised learning, we adopt a self-supervised trained network rather than ImageNet-trained networks and show it also works comparably well. Both these two discoveries are conceptually simple yet super-effective and valuable. 

\section{Method}
In the natural language processing field, the words are naturally discrete tokens which contain high semantic information. By contrast, vision signals are continuous with redundant low-level information.
While there are various ways to discretize the image in prior works, the semantic level of the resulting visual tokens has been largely ignored. 
In this section, we start by briefly describing the discrete representation learning from VQ-VAE,
and then introduce the process of how to learn a perceptual codebook,
followed by BERT pre-training over the learned perceptual visual tokens.

\begin{figure*}[t]\centering
\includegraphics[width=0.99\linewidth]{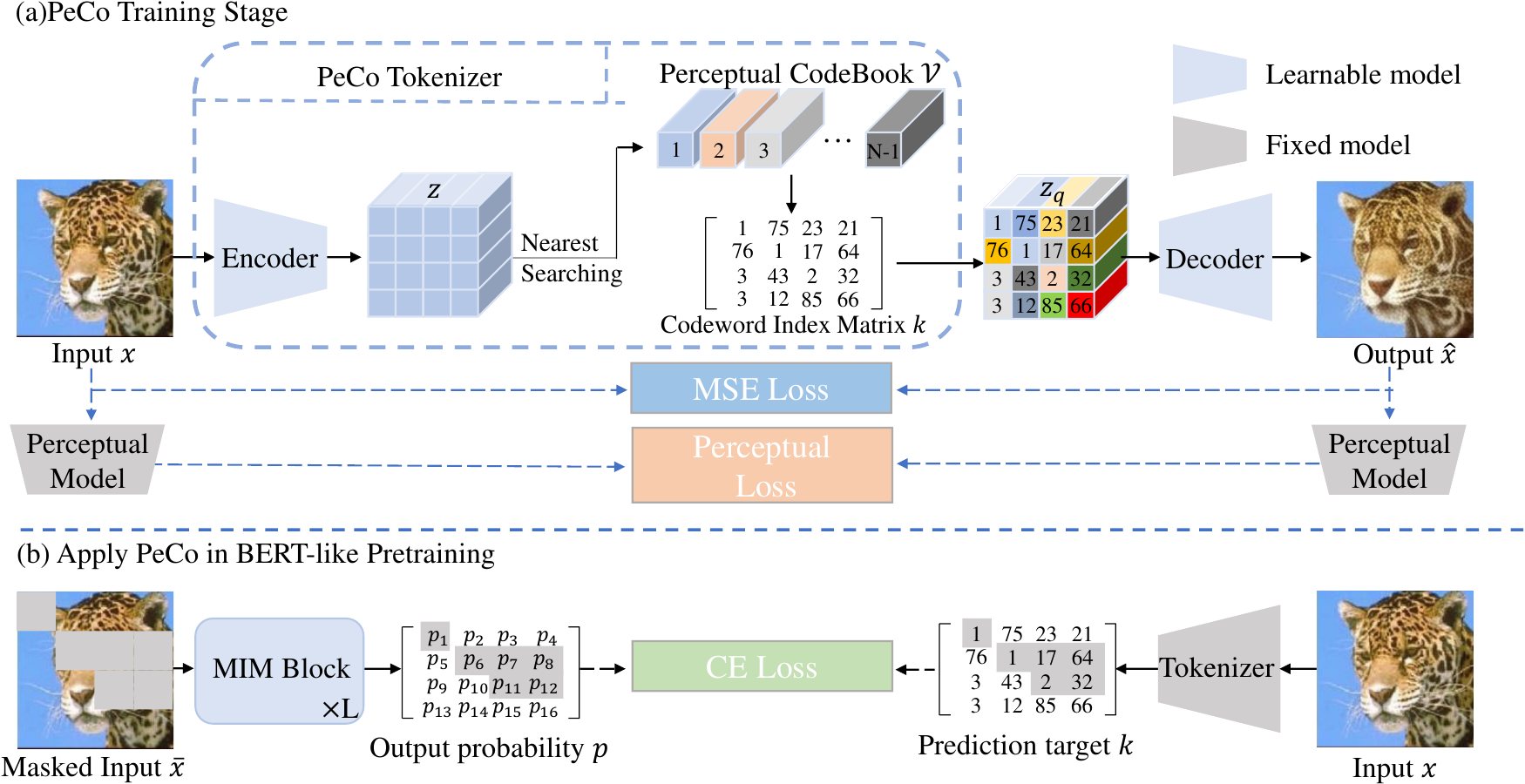}
\caption{(a) Training pipeline of our Perceptual Coodbook. (b) Apply PeCo in BERT-Like pretraining. Our PeCo provides a more semantic prediction target to the Mask Image Modeling Task. 
}
\label{fig:pipeline}
\end{figure*}

\subsection{Learning Discrete Codebook for Visual Content}
We utilize VQ-VAE~\cite{oord2017neural} 
to convert the continuous image content into the form of discrete tokens.
Consider an image $x \in \RR^{H \times W \times 3}$, VQ-VAE is able to represent it with discrete visual codewords $\{z_q^1, z_q^2, \cdots, z_q^N\} \in \mathcal{V}^1 \times \mathcal{V}^2 \times \cdots \times \mathcal{V}^N$, 
where $z_q^i$ comes from a visual codebook (vocabulary) $\mathcal{V}^i =\{e_k^i \in \mathbb{R}^D\}_{k=1}^{K_i}$ consisting of $K_i$ $D$-dimensional codewords. Usually we have $K_1 = K_2 = \cdots = K_N = K$ for simplicity,
and $N = h \times w$ with $h \times w$ being the spatial resolution of the latent space.

Specifically, to learn such latent codeooks, VQ-VAE contains three major parts: an encoder,
a quantizer and a decoder.
The encoder maps the input image to intermediate latent vectors $z = \encoder(x)$,
 where $z \in \RR^{h \times w
 	\times D}$.
 The quantizer is in charge of
quantizing each vector at position $(i,j)$ to be codewords coming from the corresponding codebook $\mathcal{V}^{i,j} = \{e_k^{i,j}\}_{k=1}^K \subset
\RR^D$ according to nearest neighbor assignment.
That is,

\begin{align}
	k^* & = q(z^{i,j}) = \argmin_{k \in \{1,2,\cdots,K\}} \Vert z^{i,j} - e_k^{i,j} \Vert. \\
	z_q^{i,j} & = r(k^*) =  e^{i,j}_{k^{*}}, 
\end{align} 
where $q$ is the quantization encoder that maps the vector to an index of the codebook,
and $r$ is the quantization decoder that reconstructs the vector from the index.
Based on the quantized codewords $\quantizedcode$, the decoder aims to reconstruct the input image $x$. Suppose the reconstruct result is $\hat{x} = \decoder(\quantizedcode)$. Since the quantizer is non-differentiable, to back-propagate gradient into encoder, the gradient is
approximated like the straight-through estimator~\cite{bengio2013estimating} and just copied from decoder to encoder~\cite{oord2017neural}. The training objective of VQ-VAE is defined as,
\begin{align}
\mathcal{L}_{\text{VQ-VAE}}(\encoder, \decoder, \{\mathcal{V}\}) &= \mathcal{L}_{pixel}
  + \Vert \text{sg}[\encoder(x)] - \quantizedcode \Vert_2^2 \nonumber \\ &+ \beta\Vert \text{sg}[\quantizedcode] - \encoder(x) \Vert_2^2.
\label{eq:origvqloss}
\end{align}
Here, $\mathcal{L}_{pixel} = \frac 1 {H\times W\times 3}\Vert x - \hat{x} \Vert$ is the per-pixel loss, $\text{sg}[\cdot]$ is the stop-gradient operator, $\beta$ is a loss weight set to 0.25 in all our experiments.

\begin{figure}[t]\centering
\includegraphics[width=0.995\linewidth]{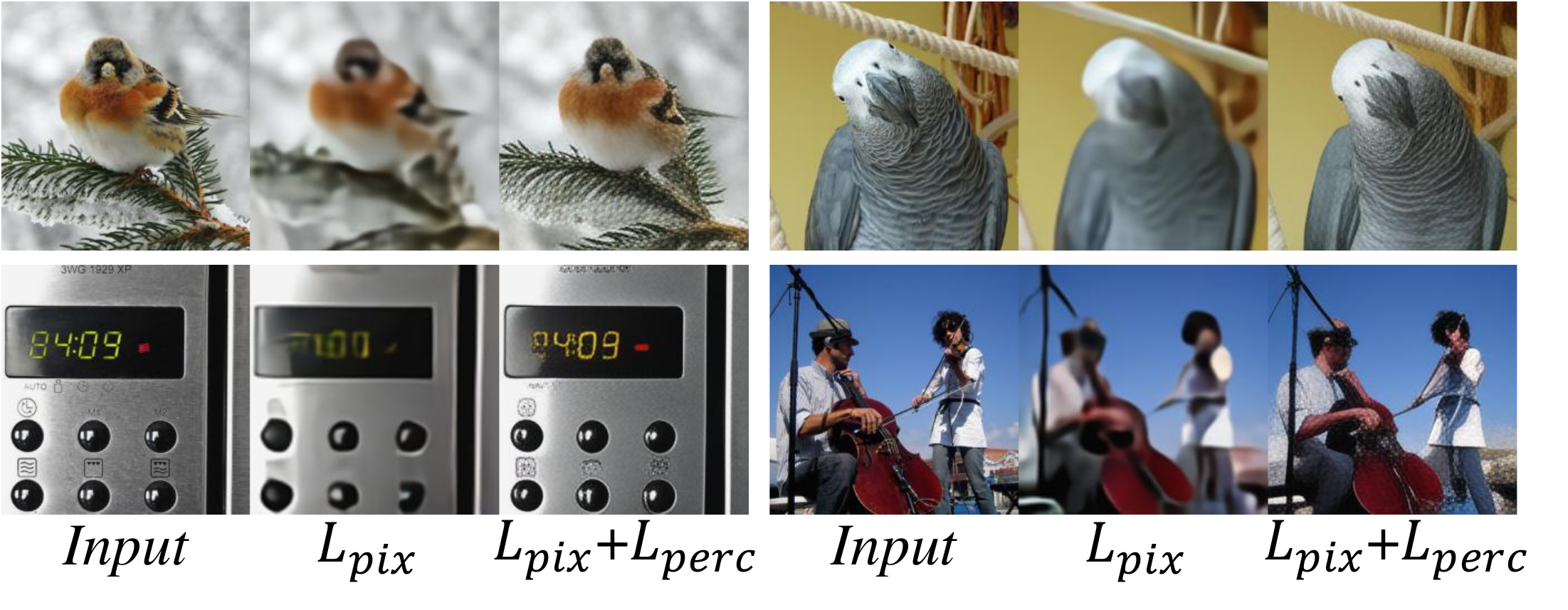}
\caption{Image reconstruction with different losses. An example contains three images showing input (left), reconstructed image using pixel-wise loss (middle), and
reconstructed image using pixel-wise and feature-wise losses (right). We can see that perceptually the right image appears more similar to the input compared with the middle image, although the middle image gets lower pixel-wise loss.}
\label{fig:recons_comparison}
\end{figure}

\subsection{Learning Perceptual Codebook for Visual Content}
In the vanilla VQ-VAE, the codebook is learned by an element-wise pixel loss, \ie $\mathcal{L}_{pixel}$, between the original image and the reconstructed image. However, this per-pixel loss may prevent the network from capturing \emph{perceptual} difference since the loss only accounts for the correctness of individual pixels. Therefore, a small shift and rotation operation on the original image may not cause perceptual change but large $\ell_1$/$\ell_2$ error.

Therefore, we propose a simple yet effective strategy by enforcing perceptual similarity between the original image and the reconstructed one beyond the pixel loss.
The perceptual similarity is not based on pixel differences but instead feature differences where the high-level image features extracted from a pre-trained deep neural network. 
We hope this feature-wise loss will better capture perceptual difference and offer invariance towards low-level variations.
We show the comparison of using different losses in Figure~\ref{fig:recons_comparison} from the perspective of image reconstruction, suggesting that images with lower pixel-wise loss may not appear perceptually similar. 

Previous works usually adopt a supervised pretrained VGG~\cite{simonyan2014very} network to calculate perceptual loss, since using supervision is not consistent with our purpose of self-supervised pre-training. We turn to the self-supervised models and replace the ConvNet-based model with Vision Transformer, which have a better modeling capability and efficiency. On the other hand, pre-trained models usually encode different levels of semantic information in different layers, to enable our codebook to have rich perceptual information, we adopt multi-scale features from multiple layers of the model to calculate the perceptual loss. Our experiments show that a vision Transformer (ViT-B model) from self-supervised learning works well for calculating perceptual loss.

Formally, let $f_l(x)$ be the normalized activations of the $l$-th layer of a network $F$ when processing the image $x$.
The size of the feature map is $H_l \times W_l \times C_l$ with $H_l$ being the height, $W_l$ being the width and $C_l$ being the channel dimension. 
Usually, multi-scale features, more comprehensive and discriminative, from multiple layers at different depth are extracted to calculate the perceptual similarity for better semantic capture. The perceptual metric for the input image $x$ and the reconstructed image $\hat{x}$ can be formulated as,
\begin{align}
\mathcal{L}_{\text{percep}} = \sum_{l \in \mathcal{S}} \frac 1 {C_lH_lW_l} \Vert f_l(x) - f_l(\hat{x}) \Vert_2^2,
\label{eq:perceptloss}
\end{align}
where $\mathcal{S}$ denotes the number of layers from which the features are extracted.

Therefore, the overall objective function is,
\begin{align}
\mathcal{L}_{\text{VQ-VAE}_{percep}} & = \mathcal{L}_{pixel} + \lambda \mathcal{L}_{percep} \nonumber \\
&+ \Vert \text{sg}[\encoder(x)] - \quantizedcode \Vert_2^2 \nonumber \\ &+ \beta\Vert \text{sg}[\quantizedcode] - \encoder(x) \Vert_2^2,
\label{eq:recloss}
\end{align}
where $\lambda$ is the hyper-parameter for the loss weight of $\mathcal{L}_{percep}$, we will study different vaules of loss weight $\lambda$ in the experiments.
The training pipeline of perceptual codebook is illustrated in Figure~\ref{fig:pipeline} (a).
After training, the encoder and the quantizer are used as tokenizer in the subsequent pre-training process.

\subsection{BERT Objective over Perceptual Codebook}
We adopt the BERT objective to perform the \emph{masked image modeling} task over the discrete  visual tokens as in BEiT~\cite{bao2021beit}, illustrated in Figure~\ref{fig:pipeline}.
For a given image $x$, 
the input tokens are image patches which are non-overlappingly split from the whole image,
and the output tokens are discrete perceptual visual words obtained through learning Eqn~\ref{eq:recloss}.
Let the input be $\{x^1, x^2, \cdots, x^N\}$, and the groundtruth output be
$\{k^1, k^2, \cdots, k^N\} = q(Enc(x))$.
The goal of the masked image modeling is to recover the corresponding visual tokens from the masked input where a portion of input tokens have been masked.

Precisely, let $\mathcal{M}$ be the set of masked index.
Then the masked input $\bar{x}$ is represented as,
\begin{align}\bar{x}^i = 
	\left\{ 
	\begin{array}{cc}
		x^i, & i \notin \mathcal{M}\\
		m, & i \in \mathcal{M}\\
	\end{array}
	\right. , i = 1,2,\cdots,N,
\end{align}
where $m$ is a learnable mask token as same dimension as non-mask tokens.
The masked input tokens are fed into a $L$-layer vision Transformer with the last layer's hidden output being denoted as $\{h^1, h^2, \cdots, h^N\}$.
We aim at recovering the corresponding visual token from the hidden vector at masked positions.
To achieve that with the classification loss, 
a $K$-way classifier is appended after the hidden vector $h^i$ to get the probability estimation about all possible discrete tokens in the corresponding codebook $\mathcal{V}^i$.
Suppose the groundtruth discrete visual tokens corresponding to the masked patches are $k^t$ with $t \in \mathcal{M}$, the pre-training objective can be formulated as,
\begin{align}
	\mathcal{L}_{\text{pre-training}} =  - \sum_{t \in \mathcal{M}} \text{log} P(k^t|\bar{x}),
	\label{eq:perceploss}
\end{align}
where $ P(k^t|\bar{x})$ is the estimated target token probability for masked patches of corrupted image $\bar{x}$.

After pre-training the model, we apply the model to various downstream tasks including ImageNet-$1$K~\cite{deng2009imagenet} classification, COCO object detection~\cite{lin2014microsoftcoco}, and ADE20K~\cite{zhou2017scene} Segmentation.

\subsection{Pre-training Details}
\noindent \textbf{Vector Quantizer.} 
We use the standard k-means algorithm for vector quantization. 
We set the codebook size $K$ as 8192 for fair comparison.
When the size of the discrete latent space $K$ is large, we observe that only a few codewords are selected to represent image and get trained.
Many other codewords are wasted.
To overcome this issue, we adopt exponential moving averages ~\cite{oord2017neural} to update the codebook which is proved to be useful for increasing utilization of codewords in a codebook. 

\noindent \textbf{Perceptual Codebook Learning Setup.} We train the perceptual codebook using the training set of ImageNet-1K dataset by default. For the encoder and decoder of VQ-VAE, we choose traditional convolutional based backbone.
The network contains two residual blocks at each resolution.
A self-attention block is applied to the smallest resolution for both encoder and decoder. 
For perceptual loss, we use the pre-trained 100 epochs ViT-B model from self-supervised method MoCo v3~\cite{chen2021empirical} by default.

\noindent \textbf{BERT Pre-training Setup.} 
For computation resource consideration, we use the original ViT-B/16~\cite{dosovitskiy2020image} as the basic architecture of our backbone to validate the effectiveness of the learned visual codebook, as in BEiT~\cite{bao2021beit}. The model is pre-trained for 300/800 epochs with the batchsize of 2048. 
We use a block-wise masking strategy for obtaining the corrupted images with the same setup as BEiT~\cite{bao2021beit}. 
We further demonstrate the effectiveness of our approach when scaling to ViT-Large and ViT-Huge backbones.

\section{Experiments}

\subsection{Downstream Tasks}

\noindent \textbf{Image Classification} aims to classify a given image into its corresponding class category. We use the popular ImageNet-1K dataset.
To enable classification,
a global average pooling layer is appended after the pre-trained model.
We finetune the model with 100 epochs and a cosine decay learning rate that warmups to $4e^{-3}$ with 20 epochs and decays to 0.
Following \cite{bao2021beit}, the layer-wise learning rate decay is also used and set to 0.65 by default. For more details, please refer to the supplementary materials.

\begin{table}[t]
\centering
\small
\resizebox{1\linewidth}{!}{
\setlength{\tabcolsep}{0.5mm}{
\begin{tabular}{lcccccc}
\toprule
\multirow{2}{*}{ Methods} & pre-train  & pre-train & \multirow{2}{*}{ViT-B} & \multirow{2}{*}{ViT-L} & \multirow{2}{*}{ViT-H} & \multirow{2}{*}{ViT-H$_{448}$} \\ & dataset & epochs &  \\

\midrule
\multicolumn{7}{l}{\textit{Training from scratch (i.e., random initialization)}} \\
ViT$_{384}$  & - & - & 77.9 & 76.5 & -- & -- \\
DeiT       & - & - & 81.8 & -- & -- & -- \\
ViT           & - & - & 82.3 & 82.6 & 83.1 & --  \\
\midrule
\multicolumn{7}{l}{\textit{Self-Supervised Pre-Training on ImageNet-1K}} \\
DINO      & IN-$1$K & 300 &  82.8 & -- & -- & -- \\
MoCo v3   & IN-$1$K & 300 &  83.2 & 84.1 & -- & --\\
BEiT            & IN-$1$K& 800 &  83.2 & 85.2 & -- & -- \\
BootMAE & IN-$1$K& 800 &  84.2 & 85.9 & -- & --\\
MAE     & IN-$1$K& \textbf{1600} & 83.6 & 85.9 & 86.9 & 87.8 \\
PeCo    & IN-$1$K& 800 & \underline{84.5} & \underline{86.5} & \underline{87.5} & \textbf{88.3} \\
\bottomrule
\end{tabular}}}
\caption{Image classification accuracy (\%) comparison on ImageNet-$1$K (IN-$1$K) of different self-supervised methods using various backbones. We report Top-1 accuracy and our method PeCo outperforms previous self-supervised methods.}
\label{tbl:cls:imagenet}
\end{table}

\noindent \textbf{Semantic Segmentation} 
is the task of assigning a label to each pixel of the input image.
We compare on the semantic segmentation dataset ADE$20K$ benchmark~\cite{zhou2017scene}.
Here we employ the Upernet~\cite{xiao2018unified} as the basic framework. For fair comparison, we follow previous works~\cite{bao2021beit} and train Upernet 160k iterations with batch size set as 16, more details are provided in the supplementary material.

\noindent \textbf{Object Detection and Segmentation.}
Object detection is to 
locate objects in a given image and identify each object.
We perform fine-tuning on the COCO objection detection and segmentation with the Mask R-CNN~\cite{he2017mask} framework. Specifically, we add four different scale FPNs to scale the feature map into different size following~\cite{bao2021beit}. The fine-tuning is conducted with ``1x'' (12 training epochs) schedule and single-scale input on the COCO training set and test the performance on COCO validation set, following the strategy used in Swin Transformer~\cite{liu2021swin}.

\begin{table}[t]
\centering
\small
\resizebox{1\linewidth}{!}{
\setlength{\tabcolsep}{1mm}{
\begin{tabular}{lcccc}
\toprule
\multirow{2}{*}{ Methods} & tokenizer  & tokenizer & BERT pre- & IN-1K
\\ & dataset & \#params & train epoch & Top-1 \\
\midrule

BEiT   & DALLE(400M) & 53.8M & 300/800 &  82.8/83.2 \\
PeCo   & IN-1K(1.3M) & 37.5M & 300/800 &  84.1/84.5 \\
PeCo$_{lite}$   & IN-1K(1.3M) & 25.7M & 300/800 &  84.0/84.5 \\

\bottomrule
\end{tabular}}}
\caption{Tokenizer comparison with BEiT. Here we report tokenizer training dataset and \#parameters. PeCo$_{lite}$ is a lite version of PeCo that reduces the channel number of tokenizer by half.}

\label{tbl:cls:tokenizer}
\end{table}

\begin{table}[t]
\centering
\small
\resizebox{1\linewidth}{!}{
\setlength{\tabcolsep}{1mm}{
\begin{tabular}{lccccc}
\toprule
\multirow{2}{*}{ Methods} & pre-train  & pre-train & ADE-20K & \multicolumn{2}{c}{COCO} 
\\ & dataset & epochs & mIoU & $\text{AP}^{\text{bb}}$ & $\text{AP}^{\text{mk}}$ \\

\midrule
DEiT & IN-1K & 300 & 47.4 & 44.1 & 39.8 \\
MoCo & IN-1K & 300 & 47.3 & 44.9 & 40.4 \\
BEiT & DALLE+IN-1K & 800 & 47.1 & 46.3 & 41.1 \\
MAE  & IN-1K & 800 & 47.6 & 46.8 & 41.9 \\
MAE  & IN-1K & 1600 & 48.1 & 47.2 & 42.0 \\
PeCo  & IN-1K & 800 & \textbf{48.5} & \textbf{47.8} & \textbf{42.6} \\

\bottomrule
\end{tabular}}}
\caption{Semantic segmentation mIoU (\%) comparison on ADE20K and object detection and instance segmentation comparison in terms of box AP ($\text{AP}^{\text{bb}}$) and mask AP ($\text{AP}^{\text{mk}}$) on COCO. The backbones for all the methods are the ViT-B.}

\label{tab:ade20coco}

\end{table}

\subsection{Comparison with previous works}
We first compare our PeCo with previous state-of-the-art works. 
Here we report ImageNet-1K results with various model sizes. For object detection on CoCo and semantic segmentation on ADE20K, we use ViT-B as the backbone.

\noindent \textbf{Image Classification.}
The Top-1 accuracy on ImageNet-1K classification is reported in Table~\ref{tbl:cls:imagenet}. 
We compare our method with 1) ViT~\cite{dosovitskiy2020image} and DeiT~\cite{touvron2021training} that are supervisedly trained from scratch with random initialization;
and 2) MoCo v3~\cite{chen2021empirical} and DINO~\cite{caron2021emerging}, represent the contrastive learning for self-supervised pre-training;
and 3) BEiT~\cite{bao2021beit}, MAE~\cite{he2021masked} and BootMAE~\cite{dong2022bootstrapped}  based on masked image modeling for self-supervised pre-training.
It can be seen that our model (PeCo) significantly improves the performance compared with the models trained from scratch, suggesting the effectiveness of pre-training.

Compared with prior self-supervised pre-training models,
our model achieves the best performance.
For example, 
our model using ViT-B backbone pre-trained with 800 epochs reaches 84.5\% Top-1 accuracy,
1.3\% higher than BEiT and 0.9\% higher than MAE.
Furthermore, we also compare the results on larger backbones, \eg ViT-L and ViT-H.
The results are reported in the Table\ref{tbl:cls:imagenet}, showing significantly better performance than previous counterparts. 
This validates that our perceptual codebook is indeed beneficial for pre-training. 
Concretely,
our model PeCo-H$_{448}$ achieves the best Top-1 accuracy, \textbf{88.3}\%, on ImageNet-1K without external data, outperforming MAE by $0.5\%$. This is a new state-of-the-art result using only ImageNet-1K data.
 
We also report the results pre-trained with 300 epochs in Table~\ref{tbl:cls:tokenizer}.
Compared with the baseline BEiT~\cite{bao2021beit}, our model gets $+1.3\%$
improvement for both 300 and 800 pre-training epochs. 
We further investigate a lite version of tokenizer which reduces the channel number of the original by half. This decreases the extra timecost introduced by the tokenizer by about $2\times$.
We can see from Table~\ref{tbl:cls:tokenizer} that with a lite tokenizer, our model still gets competitive performance.

\noindent \textbf{Semantic segmentation.}
We compare our method with 1) DEiT, which is a supervised pre-training method on ImageNet-1K , 2) MoCo, the contrastive learning based methods, and 3) BEiT~\cite{bao2021beit}, MAE~\cite{he2021masked}, the state-of-the-art self-supervised learning model. Here we use UperNet~\cite{xiao2018unified} framework with $512\times 512$ input and trained for 160K iterations. 
The evaluation metric is mean Intersection of Union (mIoU) averaged over all semantic categories and we report single-scale results here.
The results are given in Table~\ref{tab:ade20coco}. Our method achieve 48.5 mIoU, +1.1 mIoU than supervised based methods. It is also + 1.2 mIoU than MoCo, +1.4 mIoU than BEiT, and +0.9 mIoU than MAE. Our model even achieve better results(+0.4 mIoU) than MAE pre-training with 1600 epochs. This verifies the effectiveness of the perceptual codebook.

\begin{figure*}[t]\centering
\includegraphics[width=0.995\linewidth]{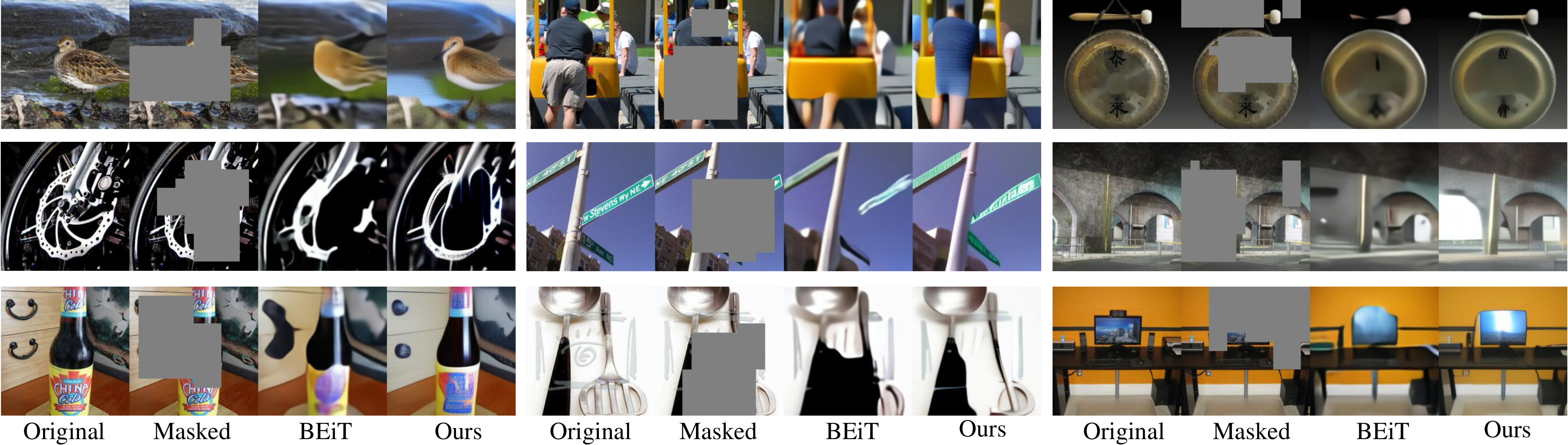}
\caption{Examples of reconstruction results on ImageNet-1K using BEiT and our PeCo.} 
\label{fig:mask_prediction_com}
\end{figure*}

\noindent \textbf{Object detection and segmentation.}
We further investigate our transfer performance on object detection and segmentation. Here we use Mask-RCNN~\cite{he2017mask} framework with single-scale input and $1\times$ schedule (12 epochs).
We compare with the strong competitor BEiT~\cite{bao2021beit} on this dataset.
The evaluation metric is box AP for detection and mask AP for segmentation.
The comparison is presented in Table~\ref{tab:ade20coco}. Our model with ViT-B as backbone achieve 47.8 box AP and 42.6 mask AP, +3.7 box AP and +2.8 mask AP over supervised methods. Our model also outperform recent work MAE by +1.0 box AP, + 0.7 box AP under the same pre-training epochs. Our model is also higher than MAE pre-training with 1600 epochs. 

\subsection{Analysis of Perceptual Codebook}
In this section, we ablate our perceptual codebook by using the setting of self-supervised pre-training on ImageNet-1K. 
The pre-raining epochs is 800.

\noindent \textbf{Semantics of the Codewords.}
The most important question would be:
\emph{will the learned perceptual codewords exhibit (more) semantic meanings?}
To answer this, 
we quantitatively evaluate the codewords' semantics from two aspects.
(1) We use the codewords of the image as features for classification. An average pooling is conducted over the quantized codewords of the image and we test its linear probing accuracy over ImageNet dataset.
(2) We use an ImageNet-1K supervisedly pre-trained DeiT-T~\cite{touvron2021training} (72.2\% Top1 accuracy on clean ImageNet val set) to test the classification accuracy over the reconstructed images.
We compare with the variant without using the perceptual similarity.
The results are given in Table~\ref{tab:vae_acc}.
We find that our perceptual codewords get much higher accuracy for both linear evaluation on codewords and classification on the reconstructed images. 
This indicates that our perceptual codebook exhibits more semantic meanings and benefits the image reconstruction process. 
We also provide a visualization of the masked region prediction using BEiT~\cite{bao2021beit} and our PeCo in Figure~\ref{fig:mask_prediction_com}, showing that our PeCo, with the aid of perceptual codebook, is able to make more semantic predictions for the masked region.


\begin{table}[t]
\centering
\footnotesize
\setlength{\tabcolsep}{4mm}{
\begin{tabular}{l|cc}
\hline
\multirow{2}{*}{ Methods} & LinearProb. & Classification.\\ 
& on codewords & on recon. \\
\hline
DALL-E  & 6.1    &  18.2 \\
PeCo(w/o $\mathcal{L}_{percep}$) & 10.2  & 17.9  \\
PeCo(ours) & 29.7   &  51.7 \\
\hline
\end{tabular}}
\caption{Evaluation of the semantics of the codewords
from linear probling accuracy (\%) of codewords on ImageNet-$1$K and classification accuracy (\%) on the reconstructed ImageNet validation images using Deit-T. }
\label{tab:vae_acc}
\end{table}
\begin{table}[t]
\centering
\setlength{\tabcolsep}{1.5mm}{
\begin{tabular}{l|c}
\hline
Loss for Tokenizer Training & acc. on IN-1K \\
\hline
$L_{pixel}$ & 82.9\\
$L_{pixel}$ + $L_{percep}$ from SSL ResNet-50 & 84.0 \\
$L_{pixel}$ + $L_{percep}$ from SSL ViT-B & 84.1 \\
$L_{pixel}$ + $L_{percep}$ from Supervised VGG & 84.1 \\
\hline
\end{tabular}}
\caption{The performance comparison when using different architectures for calculating the perceptual similarity. }
\label{tab:peco_different_percep}
\end{table}

\noindent \textbf{Deep Architectures for Perceptual Similarity.}
Another key question would be:
\emph{will the deep architectures for deep perceptual features affect the perceptual codebook learning and thus affect the pre-training performance?}
Therefore, we investigate two different deep architectures: convolutional-based backbone ResNet50~\cite{he2016deep} and Transformer-based model ViT-B~\cite{dosovitskiy2020image}.
We study the self-supervised models in order to enable unsupervised pre-training. 
The results are reported in Table~\ref{tab:peco_different_percep}.
We can see that using convolution-based or Transformer-based network achieves similar performance.
In addition, we also report the results using the classical supervised (\ie using label) trained VGG~\cite{simonyan2014very} in Table~\ref{tab:peco_different_percep}. It can be seen that using supervised model for perceptual metric achieve comparable performance as self-supervised model.

\begin{table}[t]
\centering
\small
\setlength{\tabcolsep}{4mm}{
\begin{tabular}{l|c}
\hline
Perceptual mechanism  & Top-1 acc. on IN-$1$K\\
\hline
Classification loss on codewords  & 82.9 \\
Contrastive loss on codewords  & 82.9 \\
Perceptual loss on images & 84.1 \\
\hline
\end{tabular}}
\caption{Performance comparison of our implicit way (perceptual loss on images) and the explicit ways (classification/conrastive loss on codewords) for improving the perceptual level of codebook.}
\label{tab:percep_mech}
\end{table}

\subsection{Discussions}
We present several in-depth discussions about the proposed model in this section.

\noindent \textbf{Implicit vs. Explicit.}
The key contribution of our paper is improving the perceptual level of the discrete visual tokens for the subsequent pre-training.
We have successfully demonstrated that through a simple strategy, \ie enforcing perceptual similarity over images.
One may think that it seems quite implicit for learning perceptual codebook by constraining on images instead of directly exploiting some constraint over the codebook.
Indeed, we also experiment in two explicit ways:
1) supervised classification loss over the codewords; 2) constraining a momentum contrastive loss over the quantized codewords through data augmentation in a self-supervised way. 
We hope that leveraging those forms of high-level classification objective
may encode some semantics into the codewords. 
But empirically we found that such explicit ways are not as effective as the proposed implicit strategy.
The results are reported in Table~\ref{tab:percep_mech}.
We conjecture that the codebook may learn global semantics from the classification/contrastive loss and thus fail to differentiate different codewords, which is not suitable for pre-training.
In contrast, deep features from a pre-trained deep model contain rich and dense semantics.

\begin{table}[t]
\centering
\small

\setlength{\tabcolsep}{6mm}{
\begin{tabular}{l|c}
\hline
Loss functions  & Top-1 acc. on IN-$1$K\\
\hline
$\mathcal{L}_{pixel}$ & 82.9 \\
$\mathcal{L}_{pixel}+\mathcal{L}_{percep}$  & 84.1 \\
$\mathcal{L}_{pixel}+\mathcal{L}_{percep} + \mathcal{L}_{adv}$  & 83.9\\
\hline
\end{tabular}}
\caption{Performance comparison when using different loss functions. Adding an extra adversarial loss can not bring gain to the transfer performance.}
\label{tab:adv}
\end{table}

\noindent \textbf{Perceptual Loss \emph{vs.} GAN Loss.} 
The perceptual loss is widely used in generation tasks with the goal of improving
the image quality. We ask the question that
\emph{is there a positive relation with the image quality and the perceptual level of the codebook.}
In order to explore this, we adopt another technique, adversarial loss in Generative Adversarial Nets(GANs)~\cite{goodfellow2014generative}, which has been proved to be effective in enhancing the reconstructed image.
Specifically, 
we add a patch-based discriminator D~\cite{li2016precomputed}, aiming to make the original image and the reconstructed one indistinguishable.
The adversarial loss is,
\begin{align}
	\min_{Enc,\{\mathcal{V}\},Dec}\max_{D}\mathcal{L}_{\text{adv}} = \mathrm{log} D(x) + \mathrm{log} (1 - D(\hat{x})).
	\label{eq:GAN_D_loss}
\end{align}
We add this loss with a suitable weight 0.4 to Eqn~\ref{eq:recloss}  and use the learned codebook for pre-training.
The resulting performance is shown in Table~\ref{tab:adv}.
We can see that adversarial loss can not bring gain to the transfer performance of pre-training.

\section{Conclusion}

In this paper, we argue that a good prediction target for masked image modeling should agree with human perception judgment.
Motivated by this observation, we propose a simple yet effective strategy to obtain perceptually discrete tokens, beneficial for BERT pre-training of vision transformers.
We present extensive comparisons on various downstream tasks. Our results indeed validate our hypothesis and show superior performance compared with previous state-of-the-art methods. We hope that the deep analysis about the prediction target in our work will lead to a broader exploration of this perspective and even help existing multi-modality foundation model pretraining \cite{yuan2021florence,wang2022omnivl}. 

\section*{Acknowledgements} 
This work was supported in part by the Natural Science Foundation of China
under Grant  U20B2047, 62072421, 62002334, 62102386 and 62121002.

\bibliography{aaai23}
\clearpage
\appendix
\section{More Experiments}

\noindent \textbf{Accelerated BERT pre-training over PeCo.}
Recent work MAE~\cite{he2021masked} introduces an asymmetric encoder-decoder design where the masked tokens are shifted from the encoder to the small decoder. This results in a large reduction in computation.
Inspired from this, we can also accelerate PeCo by adopting the network structure in~\cite{he2021masked} for BERT pre-training but with the proposed perceptual codebook as prediction targets.
We denote perceptual codebook using MAE framework as $\text{PeCo}_{\text{MAE}}$.
We show that $\text{PeCo}_{\text{MAE}}$ enjoys the efficiency of the framework while improving the performance through the proposed prediction target.

Here we show the results of adopting the accelerated BERT pre-training paradigm but with the proposed perceptual codebook as prediction target.
The comparison is shown in Table~\ref{tab:pecomae} for all the three downstream tasks.
We can see that our new prediction target enjoys the efficiency of the framework and also gets a higher downstream performance.

\begin{table}[h]
\centering
\small
\resizebox{1\linewidth}{!}{
\setlength{\tabcolsep}{1.5mm}{
\begin{tabular}{lccccc}
\toprule
\multirow{2}{*}{ Methods} & pre-train  & IN-1K & ADE-20K & \multicolumn{2}{c}{COCO} 
\\ & epochs & Acc. & mIoU & $\text{AP}^{\text{bb}}$ & $\text{AP}^{\text{mk}}$ \\

\midrule
MAE & 800 & 83.4 & 47.6 & 46.8 & 41.9 \\
PeCo$_{\text{MAE}}$  & 800 & 84.2 & 48.2 & 47.3 & 42.2 \\

\bottomrule
\end{tabular}}}
\caption{The effect of PeCo using accelerated BERT pre-training compared on (a) image classification, (b) semantic segmentation, and (c) object detection and instance segmentation.}
\label{tab:pecomae}
\end{table}

\noindent \textbf{Extend PeCo to Video-level Tasks.}
In our main paper, we explore PeCo on different image-level downstream tasks, here we further apply PeCo on video-level tasks. We apply PeCo to video recognition task with two wildly used dataset Kinetics-400 (K400)~\cite{kay2017kinetics} and something-something-v2 (SSv2)~\cite{goyal2017something}. We use TimeSformer and initial it with model pretrained on ImageNet-1K, and we use clips of size $8\times 224 \times 224$ and patch size is set to $16\times 16$. For PeCo and BEiT, we finetune it with 15 epochs, learning rate is set to $1.2e^{-3}$ and layer decay is 0.65. Weight decay is set to $1e^{-4}$. For supervised baseline DEiT, we set the learning rate as $1e^{-4}$ for the backbone and $1e^{-3}$ for the classification head. The batch size is set to 64 for all experiments.

As shown in Table \ref{tbl:cls:video}, our PeCo outperforms the supervised baseline DEiT and previous BEiT with a large margin, this proves the effectiveness and generalizability of PeCo.

\begin{table}[h]
\centering
\setlength{\tabcolsep}{2mm}{
\begin{tabular}{lccc}
\toprule
 Models & Pre-Train Epoch &  K400 Acc &  SSv2 Acc \\ 
\midrule
\multicolumn{4}{l}{\textit{Training from scratch (i.e., random initialization)}} \\
DEiT-B           & -- &  75.4 & 57.6\\
\midrule
\multicolumn{4}{l}{\textit{Self-Supervised Pre-Training on ImageNet-$1K$}} \\
BEiT           & 800  & 75.5 & 60.2 \\
PeCo(ours)                        & 800  &  \textbf{76.5} & \textbf{61.8}\\
\bottomrule
\end{tabular}}
\caption{Extend PeCo to video recognition task.}
\label{tbl:cls:video}
\end{table}

\noindent \textbf{The Loss Weight of Perceptual Similarity.}
In the experiments, the loss weight $\lambda$ in Eqn 5 is set as 1.
Here we present the performance under various values of $\lambda$ among 0, 0.3, 1, 3, 10.
The results are shown in Table~\ref{tab:peco_different_percep}.
We can see that using perceptual loss yields 84.1\% accuracy outperforming 82.9\% from the model without perceptual loss. 
However, further
enlarging the loss weight gets performance drop. 
One possible explanation is that large perceptual loss leads the model to pay more attention to semantic while lose some local details, while a good codebook for BERT pre-training needs both semantic and local details.

\begin{table}[h]
\centering

\setlength{\tabcolsep}{2.5mm}{
\begin{tabular}{l|ccccc}
\hline
$\lambda$    & 0    & 0.3 & 1 & 3 & 10 \\ 
ImagenNet-$1$K & 82.9 & 84.1 & 84.1 & 83.6 & 83.5 \\
\hline
\end{tabular}}
\caption{Illustrating the effect of loss weight of perceptual similarity.
We show fine-tune accuracy (\%) on ImageNet-$1$K. 
Enlarging the loss weight can not get consistent improvement, may due to the loss of local details.}

\label{tab:peco_different_percep}
\end{table}

\noindent \textbf{Adversarial Robustness Analysis.}
Here we provide analysis about the fine-tuned model adversarial robustness of different pretraining methods. Here we use two classical white-box attack method Basic Iteration Attach Method (BIM)~\cite{kurakin2016adversarial} and Momentum Iteration Attach Method (MIM)~\cite{dong2018boosting}
The attack threshold is $2/255$ and iterations are 20. 

As shown in Table \ref{tbl:cls:adversarial}, we find that compared with the vanilla DEiT, both contrastive-learning based method MoCo and mask image modeling based method BEiT and PeCo improves the adversarial robustness and PeCo performs best. While an interesting point is that only the robustness of MAE is worse than the baseline. We argue this may be because the prediction target of MAE is raw pixels (with simple pixel norm), so it pays more attention to the high-frequency of the input, which makes it sensitive to the high-frequency change of the input. On the contrary, BEiT and PeCo predicts tokens, which could be viewed as clustered and distillate target,  so the model could focus on the structure or semantic information of the input, rather than the high-frequency information.

\begin{table}[h]
\centering
\setlength{\tabcolsep}{2mm}{
\begin{tabular}{lcccc}
\toprule
 Models & Pre-Train Epoch &  Clean &  BIM & MIM\\ 

\midrule
\multicolumn{4}{l}{\textit{Training from scratch (i.e., random initialization)}} \\
DeiT-B     & -- &  81.8 & 46.2 & 50.2\\
\midrule
\multicolumn{4}{l}{\textit{Self-Supervised Pre-Training on ImageNet-$1K$}} \\
MoCo v3  & 300 & 83.2 & 51.8 & 54.9  \\
BEiT           & 300  & 82.8 & 50.2  & 53.2 \\
MAE  & 1600  & 83.6 & 37.2  & 42.2\\
PeCo(ours)         & 300  &  \textbf{84.1} & \textbf{52.5} & \textbf{55.3}\\
\bottomrule
\end{tabular}}
\caption{Adversarial robustness analysis on different self-learning methods.}
 
\label{tbl:cls:adversarial}
\end{table}

\noindent \textbf{Different Architectures for VQ-VAE.}
Here we investigate the performance when using different architectures for VQ-VAE.
We consider several variants of the network architecture. 
For encoder, we explore three models: 1) 16$\times$ down-sample encoder (our default setting); 2) 8$\times$ down sample encoder; 3) ViT-B (16$\times$ down-sample). For the 8$\times$ down-sample encoder, we remove one stage and train it with images of 112$\times$112 resolution. 
For decoder, we  use the inversed version of the corresponding decoder.
The results on ImageNet-$1$K dataset are shown in Table~\ref{tab:peco_archi}. We observe that CNN based encoders and decoders achieve better results than vision Transformer. 
We further reduce the parameters of decoder by decreasing the channel number or decreasing the depth of the network by half. 
Results shown in Table~\ref{tab:peco_archi} suggest that reducing the parameters of decoder may not hurt the fine-tuning performance of PeCo.

\begin{table}[h]
\small
\centering
\setlength{\tabcolsep}{1.5mm}{
\begin{tabular}{ll|c}
\hline
Encoder of VQ-VAE & Decoder of VQ-VAE & IN-1K Acc\\
\hline
ViT-B & ViT-B & 83.6 \\
CNN(8x) & CNN(8x) & 83.8 \\
CNN(16x) & CNN(16x) & 84.1 \\
CNN(16x) & CNN(16x) (Half Channel) & 84.0 \\
CNN(16x) & CNN(16x) (Half Depth)   & 84.1 \\

\hline
\end{tabular}}
\caption{Illustrating the effect of different architectures for training PeCo. 
CNN based encoders and decoders achieve better results than vision Transformer.}
\label{tab:peco_archi}
\end{table}

\section{Experiment Details}
In this section, we provide more detailed experimental settings about downstream tasks.

\noindent \textbf{VQ-VAE Architectures.} For convolutional encoder, the number of channels at the first stage is set to $64$, then it will be doubled in every downsample operation. we apply the Group Normalization~\cite{wu2018group} as introduced in Taming Transformer~\cite{esser2021taming}. The convolutional decoder is an inverse version of the encoder. For ViT-base encoder, we use the original structure, and use the inverse version of ViT as decoder.

\noindent \textbf{Perceptual Codebook Learning Setup.} We train the perceptual codebook using the training set of ImageNet-1K dataset by default. For the encoder and decoder of VQ-VAE, we choose traditional convolutional based backbone.
The network contains two residual blocks at each resolution.
A self-attention block is applied to the smallest resolution for both encoder and decoder. 
For perceptual loss, we use the pre-trained 100 epochs ViT-B model from self-supervised method MoCo v3~\cite{chen2021empirical} by default, and the 3rd, 6th, 9th, and 12nd layer are selected for deep features. 
We also apply the ResNet50~\cite{he2016deep} and VGG~\cite{simonyan2014very} model with the perceptual similarity calculated at the end of each stage. 
We set the perceptual loss weight $\lambda$ to 1 without special noting. Different models for providing deep features for perceptual loss are ablated in the experiments section. 
The input image size is $224 \times 224$, which is consistent with pre-training image input size, the latent codes are in a resolution of $16 \times 16$. We use EMA vector quantizer as the default quantizer algorithm. The learning rate is set $5e^{-5}$ with batchsize 128. We train the PeCo for 100 epochs and warm up the first 5000 iterations to stabilize the training process. The Adam~\cite{kingma2014adam} optimizer is used with $\beta_1$ and $\beta_2$ set to 0.5 and 0.95 respectively.

\noindent \textbf{BERT Pre-training Setup.} 
For computation resource consideration, we use the original ViT-B/16~\cite{dosovitskiy2020image} as the basic architecture of our backbone to validate the effectiveness of the learned visual codebook, as in BEiT~\cite{bao2021beit}. The model is pre-trained for 300/800 epochs with the batchsize of 2048. 
AdamW optimizer is adopted with learning rate, $\beta_1$, $\beta_2$, weight decay set to $1.5e^{-3}$, $0.9$, $0.999$, and 0.05 respectively. We also apply stochastic depth~\cite{huang2016deep} with 0.1 rate.
We use a block-wise masking strategy for obtaining the corrupted images with the same setup as BEiT~\cite{bao2021beit}. 
We further demonstrate the effectiveness of our approach when scaling to ViT-Large and ViT-Huge backbones.

\noindent \textbf{ADE20K Semantic segmentation.}
Here we use: UperNet~\cite{xiao2018unified} based on the implementation from mmsegmentaion~\cite{mmseg2020}. 
For UperNet, we follow the settings in ~\cite{bao2021beit} and use AdamW~\cite{loshchilov2017decoupled} optimizer with initial learning rate $3e^{-4}$, weight decay of 0.05 and batch size of 16 (8 GPUs with 2 images per GPU) for 160K iterations. The learning rate warmups with 1500 iterations at the beginning and decays with a linear decay strategy. We use the layer decay ~\cite{bao2021beit} for the backbone and we set it as 0.65. 
As the ViT architecture outputs features with the same size, here we add four different scale FPNs to scale the feature map into different size. Specifically, we upsample the output feature of the $4th$ block $4\times$, upsample the output feature of the $6th$ block $2\times$, keep the output feature of the $8th$ block unchanged and downsample the output feature of the $12th$ block $2\times$. 
We use the default augmentation setting in mmsegmentation including random horizontal flipping, random re-scaling (ratio range [0.5, 2.0]) and random photo-metric distortion. All the models are trained with input size $512\times512$. The stochastic depth is set to 0.2. When it comes to testing, we report single-scale test result.

\noindent \textbf{COCO Object Detection and Instance Segmentation.}
We use the classical object detection framework Mask R-CNN~\cite{he2017mask} based on the implementation from mmdetection~\cite{mmdetection}. We train it the $1\times$ schedule with single-scale input (image is resized so that the shorter side is 800 pixels, while the longer side does not exceed 1333 pixels) for 12 epochs. We use AdamW~\cite{loshchilov2017decoupled} optimizer with a learning rate of $4e^{-4}$, weight decay of 0.05 and batch size of 16. We also use the layer decay ~\cite{bao2021beit} for the backbone and we set it as 0.75. 
The learning rate declines at the $8th$ and $11th$ epoch with decay rate being 0.1. The stochastic depth is set to 0.1. 
Similar to the implementation of semantic segmentation above, we also use four different scale FPNs to scale the feature map into different size.

\section{More visual results}
\begin{figure*}[t]\centering
\includegraphics[width=0.995\linewidth]{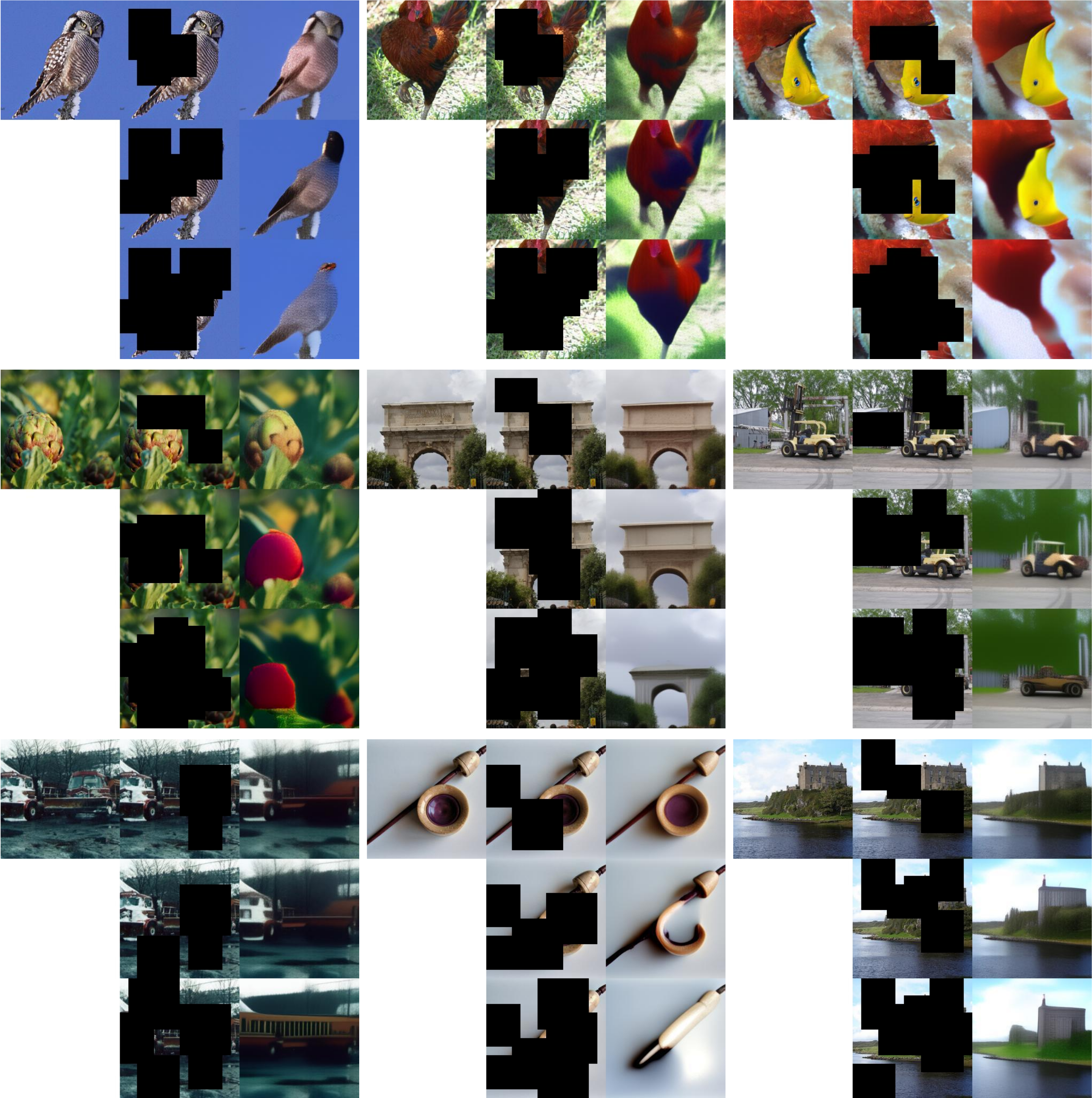}
\caption{Example samples of reconstruction task on ImageNet-$1k$ using our PeCo with different mask regions. For each sample,
the first image is the original image, the second one is the corresponding masked image, the third one is the reconstruction from perceptual codebook (PeCo). The first row masks 45 patches, the second row masks 75 patches and the lost row masks 120 patches.}
\label{fig:mask_ratio}
\end{figure*}

In Figure.\ref{fig:mask_ratio}, we show the reconstruction results with a different number of patches masked. We find that PeCo learns strong semantic that could predict a reasonable object with limited visible patches.

\end{document}